\pdfoutput=1

\documentclass[11pt]{article}

\usepackage[preprint]{acl}

\usepackage{times}
\usepackage{latexsym}
\usepackage{enumitem}

\usepackage[T1]{fontenc}

\usepackage[utf8]{inputenc}

\usepackage{microtype}

\usepackage{inconsolata}

\usepackage{graphicx}

\usepackage{multirow}
\usepackage{csvsimple}   
\usepackage{booktabs}    
\usepackage{longtable}   
\usepackage{caption}     

\usepackage{tikz}
\usepackage[edges]{forest}
\definecolor{hidden-draw}{RGB}{20,68,106}
\definecolor{hidden-pink}{RGB}{255,245,247}

\title{Multi-Party Conversational Agents: A Survey}

\author{
  Sagar Sapkota \\
  University of Central Florida \\
  \texttt{sagar.sapkota@ucf.edu}
  \And
  Mohammad Saqib Hasan \\
  Stony Brook University \\
  \texttt{mdshasan@cs.stonybrook.edu}
  \AND
  Mubarak Shah \\
  University of Central Florida \\
  \texttt{shah@crcv.ucf.edu}
  \And
  Santu Karmaker \\
  University of Central Florida \\
  \texttt{santu@ucf.edu}
}

\begin{document}
\maketitle
\begin{abstract}

\textit{Multi-party Conversational Agents} (MPCAs) are systems designed to engage in dialogue with more than two participants simultaneously. Unlike traditional two-party agents, designing MPCAs faces additional challenges due to the need to interpret both utterance semantics and social dynamics. This survey explores recent progress in MPCAs by addressing three key questions: 1) Can agents model each participants' mental states? (\textit{State of Mind Modeling}); 2) Can they properly understand the dialogue content? (\textit{Semantic Understanding}); and 3) Can they reason about and predict future conversation flow? (\textit{Agent Action Modeling}). We review methods ranging from classical machine learning to Large Language Models (LLMs) and multi-modal systems. Our analysis underscores \textit{Theory of Mind} (ToM) as essential for building intelligent MPCAs and highlights multi-modal understanding as a promising yet underexplored direction. Finally, this survey offers guidance to future researchers on developing more capable MPCAs.

\end{abstract}


\section{Introduction}
\label{sec:introduction}


\begin{figure}[!htb]
\centering
\includegraphics[width=\linewidth]{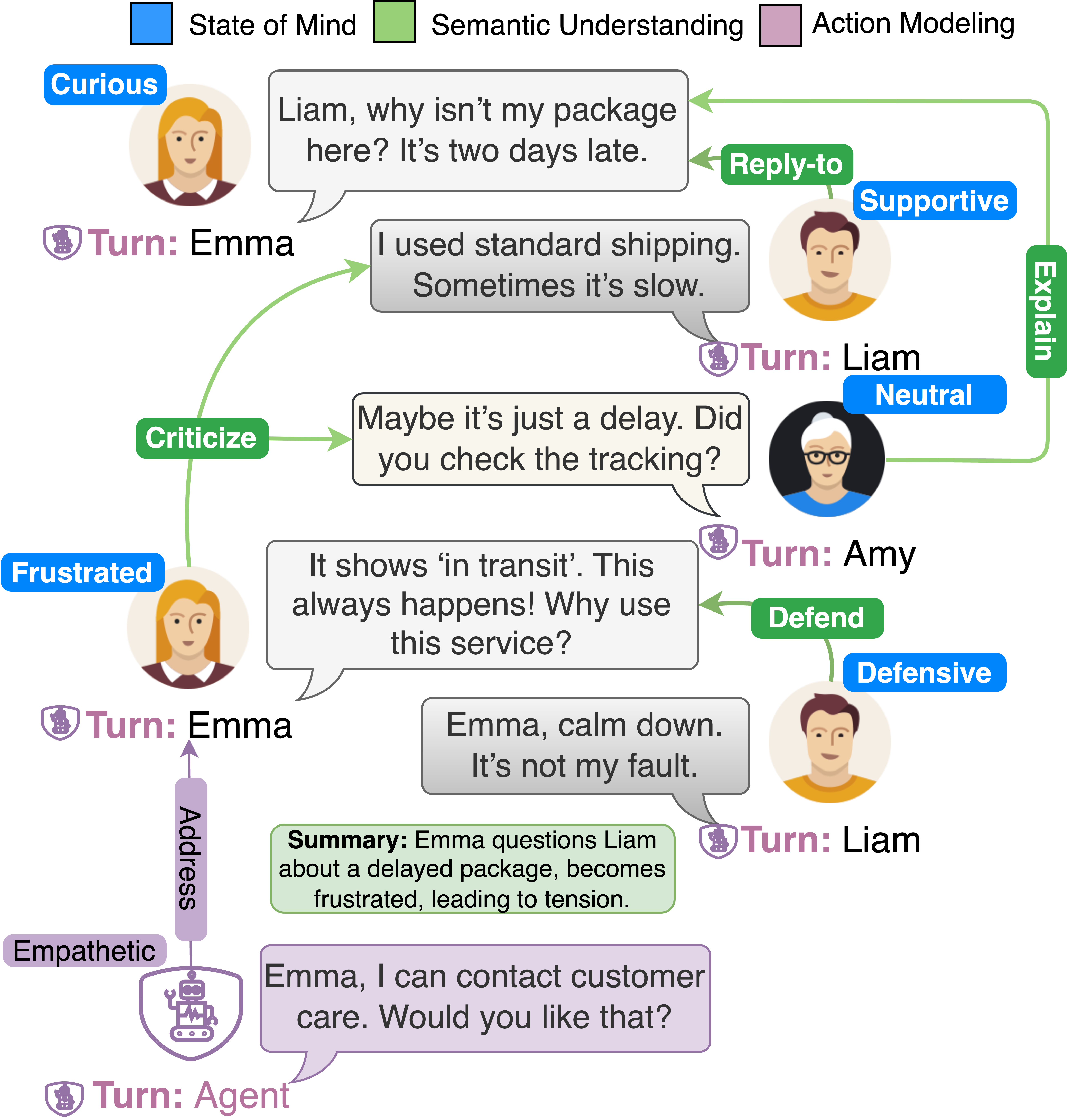}
    \caption{
    Example of a multi-party conversation demonstrating key challenges that MPCAs must handle. At each time step of the conversation, the MPCA must identify the \textcolor{blue}{states of mind} of each participant (e.g., curiosity, frustration, etc.), have \textcolor{teal}{semantic understanding} of the conversation (e.g., speaker actions like criticize and explain, dialog summary, etc.), and be able to take the appropriate \textcolor{purple}{action} (e.g., response, turn-taking, identifying addressee, etc.). Combining these capabilities makes for a social and intelligent agent.
    }
    \label{fig:mpc_example}
    \vspace{-1.5em}
\end{figure}


Developing agents for conversational understanding and natural dialogue generation has been a long-standing research agenda within the NLP community~\cite{wangInstructOnceChat2024}. Traditionally, most conversational agents have been designed for two-party interactions~\cite{Zheng_2022_ux}. However, real-world conversations—such as group chats, team meetings, online gaming, and customer support forums—often involve more than two participants. As a result, there has been growing interest in developing agents capable of engaging with multiple participants simultaneously, commonly referred to as \textbf{\textit{Multi-Party Conversational Agents}} (MPCAs).


Compared to traditional two-party systems, building MPCAs presents greater challenges due to the complex dynamics in multi-party interactions. Effective MPCAs must simultaneously infer participants' mental states~\cite{cohen2024explainablecollaborativedialogueusing}, comprehend user content~\cite{ma2023enhanced}, and anticipate future dialogue flow~\cite{houde2025controlaiagent}. Mastering these capabilities is crucial for enabling collaborative tools, socially adept robots, and intelligent assistants. For instance, Figure~\ref{fig:mpc_example} illustrates a two-speaker interaction with an agent, highlighting core challenges such as emotional variance, evolving latent states, and context-sensitive response planning~\cite{weiMultiPartyChatConversational2023a,houde2025controlaiagent}. While recent advances in conversational AI~\cite{grattafiori2024llama3herdmodels,openai2024gpt4technicalreport,qwen2025qwen25technicalreport,jiang2024mixtralexperts} have improved these components individually, integrating them into fully functional MPCAs still remains an open challenge~\cite{zhao2023chatgptequippedemotionaldialogue,tan-etal-2023-chatgpt}.

This research gap has recently attracted significant attention from the community, leading to a surge in efforts to design multi-party conversational systems~\cite{ganesh-etal-2023-survey}. In synergy with with recent advances in generative AI, particularly Large Language Models (LLMs), most of these efforts have focused on leveraging LLMs for this domain~\cite{wangMultiPartySupervisedFinetuning2024,tan-etal-2023-chatgpt}. As such, this survey aims to serve as a comprehensive resource for AI researchers interested in developing MPCAs. We review over \textbf{70} research articles relevant to multi-party conversations and categorize their contributions into three broad themes:



\smallskip
\noindent\textbf{State of Mind Modeling:} Works under this theme study the following question: ``\textit{Can models accurately infer each participant's mental and emotional states?}'' Key tasks under this theme include emotion recognition, participants' engagement detection, personality identification, and recognition of user intents.

\smallskip
\noindent{\textbf{Semantic Understanding}:} Under this theme, works assess MPCAs' ability to properly understand utterances and the context of the multi-party dialogue. This covers tasks like dialogue summarization, conversation disentanglement, discourse structure analysis, and representation learning.

\smallskip
\noindent\textbf{Agent Action Modeling:} Tasks pertaining to this theme describe MPCAs' ability to predict future conversation flow. This involves tasks like turn detection, addressee selection, and response selection/generation.

\smallskip
For each of the aforementioned themes, we start off by discussing research works utilizing traditional ML algorithms~\cite{bayser2019mlmpcdlogs, Majumder_Poria_Hazarika_Mihalcea_Gelbukh_Cambria_2019,ghosal-etal-2019-dialoguegcn} and NLP models~\cite{shen2020dialogxlallinonexlnetmultiparty, wen2024affectivenli,liDialBERTHierarchicalPreTrained2021}. We then move towards Large Language Models (LLMs) as many MPC tasks are being modeled by leveraging LLMs either through fine-tuning or prompting~\cite{wu2024silentlettersamplifyingllms,liPAPERPersonaAwareChainofThought2025,jiang-etal-2024-personallm}. Finally, since the proliferation of multi-modal data (e.g., meeting videos, group chats with audio-visual context)~\cite{zhang2024mintrec20largescalebenchmarkdataset,carletta2005ami,sasu2025akancinematicemotionsace,muller2024mpiigroupinteraction}, we discuss works where MPC tasks have been studied using visual and auditory cues~\cite{yoonBIMDRGBridgingImage2024,lee2024endofturn,jain2023multimodalperspectivesumm}. In addition, we discuss the major datasets associated with these tasks.


Based on our thematic analysis of MPCA research, we identify two promising future directions: integrating \textit{Theory of Mind} (ToM)~\cite{Premack1978} into multi-party dialogue systems and advancing \textit{Multi-modal Fusion and Grounding}. For ToM integration, we highlight gaps in the current literature and suggest tailoring models to infer mental states within task-specific contexts. For multi-modal grounding, we highlight that key tasks—such as conversation disentanglement, representation learning, addressee selection, and response generation—remain largely underexplored in multi-modal settings. Consequently, even state-of-the-art multi-modal MPCAs often struggle to simulate realistic multi-party interactions. We then discuss how incorporating open-ended multi-modal benchmarks~\cite{tang2025m3pttransformermultimodalmultiparty} can enhance MPCA development. We hope that the insights drawn from our analysis will guide the community for future research toward more robust and socially aware MPCA systems.

\section{A Thematic Taxonomy}

\tikzstyle{my-box}=[
    rectangle,
    draw=hidden-draw,
    rounded corners,
    text opacity=1,
    minimum height=1.5em,
    minimum width=5em,
    inner sep=2pt,
    align=center,
    fill opacity=.5,
    line width=0.8pt,
]
\tikzstyle{leaf}=[my-box, minimum height=1.5em,
    fill=hidden-pink!80, text=black, align=left,font=\normalsize,
    inner xsep=2pt,
    inner ysep=4pt,
    line width=0.8pt,
]
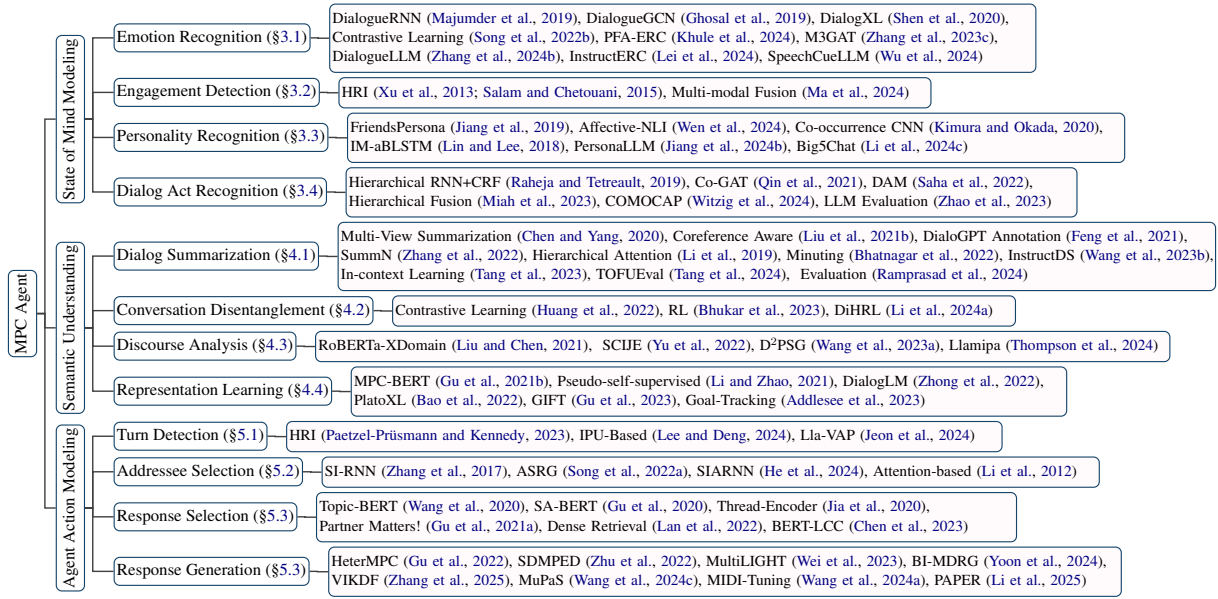
\begin{figure*}[!htb]
    \centering
    \resizebox{\textwidth}{!}{
        \begin{forest}
            forked edges,
            for tree={
                grow=east,
                reversed=false,
                anchor=base west,
                parent anchor=east,
                child anchor=west,
                base=left,
                font=\large,
                rectangle,
                draw=hidden-draw,
                rounded corners,
                align=left,
                minimum width=4em,
                edge+={darkgray, line width=1pt},
                s sep=3pt,
                inner xsep=2pt,
                inner ysep=3pt,
                line width=0.8pt,
                ver/.style={rotate=90, child anchor=north, parent anchor=south, anchor=center},
            },
            [
                MPC Agent, ver
                [
                    Agent Action Modeling, ver
                    [
                        Response Generation (\S \ref{sec:response_selection_generation})
                        [
                            HeterMPC~\cite{guHeterMPCHeterogeneousGraph2022}{, }SDMPED~\cite{zhuMultiPartyEmpatheticDialogue2022}{, }MultiLIGHT~\cite{weiMultiPartyChatConversational2023a}{, }BI-MDRG~\cite{yoonBIMDRGBridgingImage2024}{, }\\VIKDF~\cite{zhangDistillingImplicitMultimodal2025}{, }MuPaS~\cite{wangMultiPartySupervisedFinetuning2024}{, }MIDI-Tuning~\cite{wangInstructOnceChat2024}{, }PAPER~\cite{liPAPERPersonaAwareChainofThought2025}, leaf, text width=51.0em
                        ]
                    ]
                    [
                        Response Selection (\S \ref{sec:response_selection_generation})
                        [
                            Topic-BERT~\cite{wangResponseSelectionMultiParty2020}{, }SA-BERT~\cite{guSpeakerAwareBERTMultiTurn2020}{, }Thread-Encoder~\cite{jiaMultiturnResponseSelection2020}{, }\\Partner Matters!~\cite{guPartnerMattersEmpirical2021}{, }Dense Retrieval~\cite{lanExploringDenseRetrieval2022}{, }BERT-LCC~\cite{chenImprovingBERTLocal2023}, leaf, text width=45em
                        ]
                    ]
                    [
                        Addressee Selection (\S \ref{sec:addressee_selection})
                        [
                           SI-RNN~\cite{zhang2017addrespselrnn}{, }ASRG~\cite{song2022jointasrg}{, }SIARNN~\cite{he2024addrselmpccyber}{, }Attention-based~\cite{li2012attnaddresseeselect}, leaf, text width=50.0em
                        ]
                    ]
                    [
                        Turn Detection (\S \ref{sec:turn_detect})
                        [
                            HRI~\cite{paetzel2023robotturntaking}{, }IPU-Based~\cite{lee2024endofturn}{, }Lla-VAP~\cite{jeon2024llavap}, leaf, text width=46.0em
                        ]
                    ]
                ]    
                [
                    Semantic Understanding, ver
                    [
                        Representation Learning (\S \ref{sec:repr_learning})
                        [
                            MPC-BERT~\cite{gu2021mpcbert}{, }Pseudo-self-supervised~\cite{li2021selfpseudosupervisedmpc}{, }DialogLM~\cite{zhong2022dialoglm}{, }\\PlatoXL~\cite{bao2022platoxl}{, }GIFT~\cite{gu-etal-2023-gift}{, }Goal-Tracking~\cite{addlesee2023mpcgoaltrackllm}, leaf, text width=46.0em
                        ]
                    ]
                    [
                        Discourse Analysis (\S \ref{sec:discourse_structure})
                        [
                            RoBERTa-XDomain~\cite{liuImprovingMultiPartyDialogue2021}{, } SCIJE~\cite{yuSpeakerAwareDiscourseParsing2022}{, }D$^2$PSG~\cite{wangD2PSGMultiPartyDialogue2023}{, }Llamipa~\cite{thompsonLlamipaIncrementalDiscourse2024}, leaf, text width=55.0em
                        ]
                    ]
                    [
                        Conversation Disentanglement (\S \ref{sec:conv_disentangle})
                        [
                            Contrastive Learning~\cite{huangConversationDisentanglementBiLevel2022}{, }RL~\cite{bhukarEndtoEndDeepReinforcement2023}{, }DiHRL~\cite{li2024dialogdisentangle}, leaf, text width=40.0em
                        ]
                    ]
                    [
                        Dialog Summarization (\S \ref{sec:dialog_summ})
                        [
                            Multi-View Summarization~\cite{chen2020multiviewseq2seqds}{, }Coreference Aware~\cite{liu2021corefawareds}{, }DialoGPT Annotation~\cite{feng2021dilogptannotatedsummarizer}{, }\\SummN~\cite{zhang2022summn}{, }Hierarchical Attention~\cite{li_keep_on_topic_2019}{, }Minuting~\cite{bhatnagar2022multilingualmpcminute}{, }InstructDS~\cite{wang2023instructds}{, }\\In-context Learning~\cite{tang2023iclllmdialogsumm}{, }TOFUEval~\cite{tang2024tofueval}{, } Evaluation~\cite{ramprasad2024llmbehavinds}, leaf, text width=56.5em
                        ]
                    ]
                ]
                [
                    State of Mind Modeling, ver
                    [
                        Dialog Act Recognition (\S \ref{sec:dar})
                        [
                            Hierarchical RNN+CRF~\cite{raheja-tetreault-2019-dialogue}{, }Co-GAT~\cite{Qin_Li_Che_Ni_Liu_2021}{, }DAM~\cite{saha-etal-2022-meta}{, }\\Hierarchical Fusion~\cite{miah-etal-2023-hierarchical}{, }COMOCAP~\cite{witzig2024multimodaldac}{, }LLM Evaluation~\cite{zhao2023chatgptequippedemotionaldialogue}, leaf, text width=47.0em
                        ]
                    ]
                    [
                        Personality Recognition (\S \ref{sec:personality_recognition})
                        [
                            FriendsPersona~\cite{jiang2019automatictextbasedpersonalityrecognition}{, }Affective-NLI~\cite{wen2024affectivenli}{, }Co-occurrence CNN~\cite{kimura2020cooccurrencecnn}{, }\\IM-aBLSTM~\cite{lin2018interlocutormodattentblstm}{, }PersonaLLM~\cite{jiang-etal-2024-personallm}{, }Big5Chat~\cite{li2024big5chatshapingllmpersonalities}, leaf, text width=50.0em
                        ]
                    ]
                    [
                        Engagement Detection (\S \ref{sec:engagement_detection})
                        [
                            HRI~\cite{xuDesigningEngagementawareAgents2013,salamEngagementDetectionBased2015}{, }Multi-modal Fusion~\cite{maMultimodalFusionLLMs2024}, leaf, text width=38.0em
                        ]
                    ]
                    [
                        Emotion Recognition (\S \ref{sec:emotion_recognition})
                        [
                            DialogueRNN~\cite{Majumder_Poria_Hazarika_Mihalcea_Gelbukh_Cambria_2019}{, }DialogueGCN~\cite{ghosal-etal-2019-dialoguegcn}{, }DialogXL~\cite{shen2020dialogxlallinonexlnetmultiparty}{, }\\Contrastive Learning~\cite{song-etal-2022-supervised}{, }PFA-ERC~\cite{khule-etal-2024-pfa}{, }M3GAT~\cite{zhang2023m3gat}{, }\\DialogueLLM~\cite{zhang2024dialoguellmcontextemotionknowledgetuned}{, }InstructERC~\cite{lei2024instructercreformingemotionrecognition}{, }SpeechCueLLM~\cite{wu2024silentlettersamplifyingllms}, leaf, text width=45.4em
                        ]
                    ]
                ]  
            ]
        \end{forest}
    }
    \vspace{-6mm}
    \caption{Thematic taxonomy of MPC tasks and recent works focusing on these tasks.
    }
    \label{fig:taxo_of_mpc_tasks}
    \vspace{-2mm}
\end{figure*}

To organize this survey paper, we create a taxonomy of existing research efforts on multi-party conversations (MPC) by categorizing them into three core themes: \textit{State of Mind Modeling}, \textit{Semantic Understanding}, and \textit{Agent Action Modeling}. This taxonomy reflects essential capabilities for human-like social communication in group settings. Indeed, a socially intelligent agent must determine when to speak, whom to address, and what to say, collectively referred to as \textit{actions}. According to Social Identity Theory~\cite{tajfel_turner_1979}, such actions are influenced by the agent’s own and others’ mental states, while reciprocal effects also exist, where actions can shape mental states~\cite{hatfield1993emotional}. Semantic understanding~\cite{stefan2021} further mediates these dynamics by enabling agents to interpret social cues and context~\cite{Rijpma2023}. As illustrated in Figure~\ref{fig:mpc_example}, these three components interact dynamically throughout the course of a conversation.

Next, we formalize our proposed taxonomy by mapping existing MPC tasks to these three themes as follows: tasks aligned with \textit{State of Mind Modeling} include emotion recognition, personality detection, dialog act recognition, and engagement detection; tasks supporting \textit{Semantic Understanding} include dialog summarization, conversation disentanglement, discourse analysis, and representation learning; and tasks contributing to \textit{Agent Action Modeling} include turn detection, addressee selection, and response generation. A visual taxonomy of these skills and associated tasks is shown in Figure~\ref{fig:taxo_of_mpc_tasks}, with an extended taxonomy—including technique categorization (traditional, multi-modal, and LLM-based)—available in Appendix~\ref{sec:mpc_task_taxonomy_full}.

\section{Research on State of Mind Modeling}
\label{sec:state_of_mind}

Understanding the state of mind of participants is essential for developing socially intelligent multi-party conversational agents.
But what exactly constitutes a participant’s \textit{state of mind}? Our survey identifies four key components that together provide a comprehensive understanding of participants' mental states during conversation: emotion recognition, engagement detection, personality recognition, and conversational intentions (i.e., dialog act recognition). These essential components of mental states are defined and modeled by legacy psychological frameworks, such as the OCEAN model for personality traits~\citep{kasap_making_2009} and the OCC model~\cite{Ortony_Clore_Collins_1988} for emotions~\cite{irfan_dynamic_2020,gebhard2005alma}. Below, we discuss these four key components pivotal to \textit{State of Mind Modeling}.

\subsection{Emotion Recognition}
\label{sec:emotion_recognition}
Emotion recognition (ER) is the task of classifying emotions as positive (e.g., joy) or negative (e.g., anger)~\cite{Ortony_Clore_Collins_1988}. It helps MPCAs to perceive and respond empathetically. 
Initial approaches did contextual modeling~\cite{sun-etal-2021-discourse-aware,Xie2024MultiViewCP} using models like recurrent neural network-based DialogueRNN~\cite{Majumder_Poria_Hazarika_Mihalcea_Gelbukh_Cambria_2019} and graph convolutional neural network-based DialogueGCN~\cite{ghosal-etal-2019-dialoguegcn} that focus on the speaker's emotional context. 
Other works tried to solve different limitations of ER:~\citet{shen2020dialogxlallinonexlnetmultiparty} used XLNet~\cite{yang2020xlnetgeneralizedautoregressivepretraining} to handle long-term emotional dependencies through integrating global and speaker attentions, while~\citet{Li_Yan_Qiu_2022,song-etal-2022-supervised} used contrastive learning to help models differentiate similar emotional signals. Other methods include isolating emotion contexts by incorporating commonsense knowledge, speaker relationships~\cite{xu2023miei}, and conversation disentanglement~\cite{zhao-etal-2022-mucdn}; assessing emotional changes using personality~\cite{wang-etal-2024-emotion} and pseudo-future forecasting~\cite{khule-etal-2024-pfa}.

Recently, LLMs and multi-modal models are common for these tasks. Multi-modal methods work by fusing text, audio, and image via attention and/or loss functions~\cite{chudasama2022m2fnet, rasendrasoa2022realtimemultimodalerc}, showing how integrating facial expressions~\cite{zheng-etal-2023-facial, zhang2023m3gat} and topic modeling~\cite{yuan2024tgmfn} enhance ER performance. LLM-based methods like structured prompting and fine-tuning~\cite{tan-etal-2023-chatgpt,zhao2023chatgptequippedemotionaldialogue} have been shown to improve ER performance significantly. Techniques like multi-step prompting and generative tasks better captured emotional dynamics~\cite{hama2024ercmultistepprompt, lei2024instructercreformingemotionrecognition}, while adding vocal descriptions bridged modality gaps~\cite{wu2024silentlettersamplifyingllms}.

\subsection{Engagement Detection}
\label{sec:engagement_detection}

Engagement detection measures the interaction level of participants in conversations, essential for managing interactions and turns between participants.
Despite being a difficult problem, recent multi-modal methods used visual and audio cues, enhancing performance by continuously tracking participants' behaviors~\citep{bohusModelsMultipartyEngagement2009,xuDesigningEngagementawareAgents2013}. Other methods increase accuracy by integrating group interaction and spontaneous conversation~\citep{salamEngagementDetectionBased2015,fedotovMultimodalApproachEngagement2018}.
Recently, multi-modal LLMs have shown promise, offering interpretability and flexibility for engagement detection~\citep{maMultimodalFusionLLMs2024}. 
However, effectively using multi-modal frameworks and LLMs for this task remains a challenge.


\subsection{Personality Recognition}
\label{sec:personality_recognition}

Personality recognition (PR) helps MPCAs identify personality traits of participants from conversation modalities (video, text, etc.), commonly using the Big-Five personality model: extraversion, agreeableness, conscientiousness, neuroticism, and openness~\cite{cunningham1977personalitystructure}. 
Initial text-based methods analyzed and aggregated predictions over dialogues to predict personality~\cite{rissola2019prccapsnet}. Later, transformers like BERT and RoBERTa enhanced PR through attention mechanisms~\cite{jiang2019automatictextbasedpersonalityrecognition}. Other approaches framed PR as a natural language inference task using emotion annotations~\cite{wen2024affectivenli}.

Multi-modal methods for PR integrate vision, audio, and behavioral cues to solve the task. 
Initial methods clustered co-occurring non-verbal behaviors~\cite{okada2015personalityclassification}, while advanced techniques employ convolutional neural networks (CNNs)~\cite{kimura2020cooccurrencecnn} and attention-based models~\cite{lin2018interlocutormodattentblstm}.
Recent multi-modal advancements combine information from wearable devices and video data~\cite{quiros2022personalitywearabledevcam}. 
Facial expressions and speech overlap are also important multi-modality features for models to do effective PR~\cite{song2023personspecificcognition,yu2019identifyingpersonalitytraitsusing}. With the widespread use of LLMs, recent progress involves shifting focus from doing PR to incorporating personalities in language models~\cite{jiang-etal-2024-personallm}. In this paradigm, tuning LLMs on target personality datasets is more effective than using general prompt-based methods~\cite{li2024big5chatshapingllmpersonalities}.

\subsection{Dialog Act Recognition}
\label{sec:dar}

Dialog Act Recognition (DAR) classifies participant utterances into certain actions, such as whether they are asking questions or making statements.
Different methods have been applied to improve upon this task: hierarchical CNNs and RNNs, which model current utterance DAR based on previous predictions for better context~\cite{liu-etal-2017-using-context}; self-attention and Conditional Random Fields (CRFs) to capture utterance and conversation level context~\cite{raheja-tetreault-2019-dialogue}; joint modeling of dialogue act and sentiment through co-attention and Graph Attention Networks (GAT)~\cite{Qin_Che_Li_Ni_Liu_2020, Qin_Li_Che_Ni_Liu_2021}; and speaker transitions and pretrained embeddings to incorporate speaker information~\cite{he-etal-2021-speaker-turn}.

Recently, multi-modal approaches integrating non-verbal cues, such as tone and facial expressions, via attention-based multi-modal transformers further improved DAR~\cite{saha-etal-2020-towards, saha-etal-2022-meta, witzig2024multimodaldac}. 
Hierarchical multi-modal fusion using early-stage Bi-LSTMs and late-stage CNN-based encoders was also shown to be effective~\cite{miah-etal-2023-hierarchical}. LLMs like ChatGPT, however, were less effective for DAR tasks~\cite{zhao2023chatgptequippedemotionaldialogue}.
Methods like dual-process masking, emphasizing informative tokens, showed improvements across LLMs like BERT, RoBERTa, and Llama~\cite{kim-etal-2024-dual}. However, LLM-based DAR approaches are still an unsolved challenge.

\section{Research on Semantic Understanding}
\label{sec:semantic}

Proper semantic understanding is essential for any conversational system. 
But what comprises \textit{semantic understanding}? 
We identify four tasks that collectively define semantic understanding in MPC: \textit{Conversation disentanglement}~\cite{elsner2008you}, \textit{Dialogue summarization}~\cite{zhu2006summarization}, \textit{Discourse Structure Analysis}~\cite{WEBBER_EGG_KORDONI_2012}, and \textit{Representation Learning}~\cite{lowe2015ubuntu}. 
These tasks help MPCAs interpret the meaning and flow of conversation, which is challenging due to overlapping utterances~\cite{sugawara2012interactive}, parallel threads of discussion~\cite{mayfield2012hierarchical}, temporal dependencies~\cite{xing2023relational}, and dynamic context shifts~\cite{galley2003discourse}.
Traditionally, graph-based and sequential models were used~\cite{wangStructureSelfAwareModel2021,gu-etal-2023-gift,shiDeepSequentialModel2019}, but recent works shifted towards transformer-based architectures~\cite{xie2024unimpc,ma2023enhanced}. In the following sections, we review existing works on each aspect of semantic understanding.

\subsection{Dialog Summarization}
\label{sec:dialog_summ}

Dialog summarization (DS) involves creating coherent summaries of conversations, highlighting the main topics and participant actions.
This task can be challenging due to overlapping topics, informal language, and varied speaker roles.
As a result, methods use multiple views like global structure, utterance-level details, topic segmentation, dialogue progression, and coreference resolution for coherent summaries~\cite{chen2020multiviewseq2seqds,liu2021corefawareds, feng2021dilogptannotatedsummarizer}. For long dialogues, works propose using retrieve-then-summarize or recursive segmentation (split-then-summarize) or segment-wise summaries for better summarization~\cite{zhang2021long_dialog_summ, zhang2022summn, han2024tgds, yin2024twostagellmdialogsumm}.

In recent times, dialogue summaries have incorporated multi-modal methods, as leveraging audio and visual cues improves topic relevance and speaker importance estimation~\cite{li_keep_on_topic_2019}. Technologies like automatic speech recognition, machine translation, perspective-aware encoding, and topic modeling have also been developed for this task~\cite{bhatnagar2022multilingualmpcminute, jain2023multimodalperspectivesumm}. LLMs are also being used for dialogue summarization but are not as effective due to hallucination, as current models, including GPT-4, continue to generate plausible but unsupported details~\cite{tang2023iclllmdialogsumm,tang2024tofueval,ramprasad2024llmbehavinds}. Works like~\citet{xu2022narratedialog} mitigate this by formatting dialogues to pre-training formats, while~\citet{zhu2024factualdialogsummfromllm} employ symbolic knowledge distillation and contrastive learning to enhance the factual consistency. LLMs are also being used to generate synthetic data for the summarization task~\cite{mishra2023llmaidedextractsumm, wang2023instructds}.

\subsection{Conversation Disentanglement}
\label{sec:conv_disentangle}

Conversation disentanglement is the task of classifying overlapping utterances into coherent threads.
Traditionally, hierarchical neural models and clustering algorithms were used but failed to capture global context and required ample data~\cite{jiangLearningDisentangleInterleaved2018,liDialBERTHierarchicalPreTrained2021}.
This problem was mitigated by contrastive learning via clustering related utterances without the need for large datasets~\cite{huangConversationDisentanglementBiLevel2022,gaoCluCDDContrastiveDialogue2023}, unsupervised learning via self-generated pseudo labels~\cite{liuUnsupervisedConversationDisentanglement2021}, and reinforcement learning by directly optimizing on coherence using reward functions~\cite{bhukarEndtoEndDeepReinforcement2023}. 
Recently, heterogeneous discourse graphs combined with Graph Convolutional Networks (GCN) have outperformed GPT-4, showing that adding structured discourse can help with this task~\cite{li2024dialogdisentangle}.


\subsection{Discourse Structure Analysis}
\label{sec:discourse_structure}

Discourse-structure analysis predicts dependencies among elementary discourse units (EDUs), such as speakers and utterances, aiding MPCAs in contextual understanding. Major approaches for discourse structure analysis include: sequential decision models that are prone to long-distance errors and error-propagation~\cite{shiDeepSequentialModel2019}; graph-based methods with EDU graphs and gated message passing~\cite{wangStructureSelfAwareModel2021}; utilizing structure distillation and speaker-aware masks to refine interaction parsing~\cite{yuSpeakerAwareDiscourseParsing2022,ma2023enhanced}; dialogue-based pretraining for improved robustness~\cite{liuImprovingMultiPartyDialogue2021}; and joint optimization with auxiliary tasks like QA~\cite{heMultitaskingDialogueComprehension2021}.

Recent advances also incorporate multi-modal inputs (vision, audio, text) via cross-modal attention~\cite{gongMODDPMultimodalOpendomain2024} or reframe parsing as a Seq2Seq task using LLMs~\cite{wangD2PSGMultiPartyDialogue2023}. LLMs also enable incremental discourse graph generation~\cite{thompsonLlamipaIncrementalDiscourse2024} and explanation-driven supervision to train smaller models in distinguishing correct structures for better accuracy~\cite{liuEnhancingMultipartyDialogue2025}.

\subsection{Representation Learning}
\label{sec:repr_learning}

Representation learning aims to encode elements of the conversation, like utterances and speaker roles, into a vector latent space. Most works employ label-agnostic self-supervised methods with custom objectives like reply-to prediction, speaker identification, and key-utterance detection for representation learning~\cite{gu2021mpcbert,li2021selfpseudosupervisedmpc,zhong2022dialoglm} with improvements like role embeddings or speaker-specific attention~\cite{bao2022platoxl,ma2023enhanced}. Another label-agnostic method is inferring hidden dialogue structures (e.g., speaker roles, utterance connections) using expectation-maximization~\cite{li2023empretainingmpdrg,li2023latentdiscourseinf}.
Also, methods where dialogues are modeled as graphs tend to enhance transformer-based representation~\cite{gu-etal-2023-gift} and adapt to tasks like emotion recognition or dialogue act~\cite{xie2024unimpc}.

Despite advances in dialogue encoding, multi-modal representation learning is still behind, mainly due to data scarcity. LLM techniques like zero-shot and few-shot prompting fail without fine-tuning in complex dialogues, lacking explicit modeling of discourse structures or speaker roles, limiting LLMs' effectiveness as encoders~\cite{addlesee2023mpcgoaltrackllm}.

\section{Research on Agent Action Modeling}
\label{sec:agent_action}
Agent action modeling predicts the next move of participants, such as deciding turn-taking, whom to address, and selecting and generating appropriate responses, for a smooth and coherent conversation.
To do this, MPCAs must decide when (turn detection), to whom (addressee selection), and what to speak (agent response).
These abilities make MPCAs natural and effective for interaction, which are discussed further below.


\subsection{Turn Detection}
\label{sec:turn_detect}

Turn detection involves knowing when a speaker ends their utterance and who should start next and is essential for ensuring smooth speaker transitions.
Simple techniques like maximum likelihood estimation, support vector machines, and convolutional neural networks failed in this task due to subtle turn shifts~\cite{bayser2019mlmpcdlogs}, so annotations with syntax and prosodic features were introduced~\cite{enomoto2020analyticturnannot}. Sequence-based neural models like GRUs and transformers further advanced turn prediction~\cite{lee2024endofturn}.

Overcoming text-only methods, multi-modality integrated speech pauses, gaze, and listener behaviors~\cite{zarkowski2019mpxturntakinghri,paetzel2023robotturntaking,lee2023turnanalysismpc}. 
LLM-based methods, like multi-task instruction tuning and prompting, enabled more flexible cue interpretation for detecting turns and backchannels~\cite{wang2024turnacousticllm,pinto2024predictiveturn}. 
Recent innovations also include full-duplex models to handle overlapping speech and real-time feedback~\cite{veluri2024syncllmduplexagents}, strategies using social reasoning~\cite{nonomura2024whospeaksnext}, and hybrid LLM-VAP (voice activity detection) methods effective across fluent dialogue~\cite{jeon2024llavap}.

\subsection{Addressee Selection}
\label{sec:addressee_selection}

Addressee selection identifies intended listeners. Early methods used static embeddings or recurrent neural networks (RNNs) to model roles and context~\cite{ouchi2016addresseerespselect}. 
Later approaches dynamically updated embeddings by jointly optimizing addressee and response selection for coherence~\cite{zhang2017addrespselrnn}. 
For low-resource languages, multilingual embeddings, adversarial learning, and transfer learning were used to enhance performance~\cite{sato2018multilingualaddrsel}.
Recent models combine addressee selection with response generation using attention mechanisms and role-aware transformers~\cite{song2022jointasrg,he2024addrselmpccyber}.
Further, multi-modal data is used to assess engagement through gaze, posture, and lip movements to compute attention scores~\cite{li2012attnaddresseeselect}.




\begin{table*}[!htb]\large
    \centering
    \resizebox{1.00\textwidth}{!}{%
    \begin{tabular}{l|c|l|c|c|c}
    \hline
        \textbf{Dataset} & \textbf{Modality} & \textbf{Model} & \textbf{Task} & \textbf{Eval Metric} & \textbf{Eval Result} \\ \hline\hline
        EmoryNLP \cite{zahiri2017emotiondetectiontvtranscripts} & Text & PFA-ERC \cite{khule-etal-2024-pfa} & Emotion Recognition & F1 Score & 42.94 \\ \hline
        MELD \cite{poria-etal-2019-meld} & Audio, Video, Text & DialogueLLM \cite{zhang2024dialoguellmcontextemotionknowledgetuned} & Emotion Recognition & F1 Score & 71.9 \\ \hline
        EmoryNLP \cite{zahiri2017emotiondetectiontvtranscripts} & Text & Affect-NLI \cite{wen2024affectivenli} & Personality Recognition & Accuracy & 61.3 \\ \hline
        DailyDialog \cite{li-etal-2017-dailydialog} & Text & COMOCAP \cite{witzig2024multimodaldac} & Dialog Act & F1 Score & 79.66 \\ \hline
        EMOTyDA \cite{saha-etal-2020-towards} & Audio, Video, Text & DAM (Text+Video) \cite{saha-etal-2022-meta} & Dialog Act & F1 Score & 57.01 \\ \hline
        SAMSum \cite{gliwa-etal-2019-samsum} & Text & InstructDS \cite{wang2023instructds} & Dialog Summary & ROUGE-1 & 58.4 \\ \hline
        Ubuntu IRC \cite{kummerfeld-etal-2019-large} & Text & Bi-Level CL \cite{huangConversationDisentanglementBiLevel2022} & Disentanglement & F1 Score & 70.4 \\ \hline
        STAC \cite{asher-etal-2016-discourse} & Text & D2PSG \cite{wangD2PSGMultiPartyDialogue2023} & Discourse Parsing & F1 Score & 62.77 \\ \hline
        CCPE \cite{radlinski-etal-2019-coached} & Text & LLa-VAP \cite{jeon2024llavap} & Turn Detection & F1 Score & 83.13 \\ \hline
        Ubuntu IRC \cite{kummerfeld-etal-2019-large} & Text & ASRG \cite{song2022jointasrg} & Addressee Selection & Accuracy & 84.65 \\ \hline
        Ubuntu IRC \cite{kummerfeld-etal-2019-large} & Text & Dense Retrieval \cite{lanExploringDenseRetrieval2022} & Response Selection & Recall & 91 \\ \hline
        TopDial \cite{wang-etal-2023-target} & Text & MIDI-Tuning \cite{wangInstructOnceChat2024} & Response Generation & BLEU & 39.6 \\ \hline
    \end{tabular}
    }\vspace{-1mm}
    \caption{Benchmark datasets for multi-party conversations and performance of the state-of-the-art models on them.}
  \label{tab:datasets_results}
  \vspace{-1mm}
\end{table*}
\subsection{Agent Response}
\label{sec:response_selection_generation}
Agent response skills help MPCAs deliver contextually appropriate and coherent utterances.
There are two ways MPCAs can respond: response selection and response generation.
Response selection involves selecting the best candidate out of a set of predefined responses. There are different methods for effective response selection: discourse parsing to separate mixed conversations for improved context matching~\cite{wangResponseSelectionMultiParty2020,jiaMultiturnResponseSelection2020,zhangSAMMultiturnResponse2023}; using speaker identity and roles~\cite{guSpeakerAwareBERTMultiTurn2020}; persona-aware models using learned persona embeddings~\cite{guPartnerMattersEmpirical2021,liDialogueHistoryMatters2021}; and graphs linking persona with dialogue elements for responses relevant to participants' preferences~\cite{juLearningImprovePersona2022}. Furthermore, there are works implementing contrastive learning~\cite{liSmallChangesMake2021,fengLearningMultiturnResponse2023}, dense retrieval~\cite{lanExploringDenseRetrieval2022}, and local context predictions~\cite{chenImprovingBERTLocal2023,suBERTKRSBERTBasedModel2024} for selecting better context-response pairs.

Response generation, on the other hand, synthesizes the response, ensuring coherence across speakers and conversation threads.
Traditional approaches used graphs capturing semantics between dialog and speaker to generate responses~\cite{guHeterMPCHeterogeneousGraph2022, juLearningImprovePersona2022}. Adding emotional dynamics~\cite{zhuMultiPartyEmpatheticDialogue2022}, speaker persona~\cite{chauhanMHaDiGMultilingualHumoraided2023}, and discourse relations~\cite{liChatMDGDiscourseParsing2024, fanImprovingMultipartyDialogue2024} within these graphs also enhanced empathy, engagement, and conversational flow, leading to better responses. Multi-modal methods introduced visual grounding ~\cite{chenMultimodalIncrementalTransformer2021, sunMultimodalDialogueResponse2022, luoForwardCreationReverse2023} and employed cross-modal attention with contrastive learning to synchronize multi-modal information~\cite{zhangZRIGFInnovativeMultimodal2023, yoonBIMDRGBridgingImage2024, zhangDistillingImplicitMultimodal2025} for improved generation of responses. Alternatively, LLMs brought new opportunities for response generation by leveraging their pretraining to learn speaker roles, turn-taking, conversational threads, and persona dynamics~\cite{wangMultiPartySupervisedFinetuning2024, wangInstructOnceChat2024, liPAPERPersonaAwareChainofThought2025, tThreadDetectionResponse2024a}. Some research also extends LLM response generation to multilingual settings~\cite{gunsonHolisticEvaluation2024, huCanXLLMsUnderstand2025}.

\section{Evaluation and State-of-the-Art}
\label{sec:eval_datasets}

So far, we have discussed a range of tasks within our thematic taxonomy, each associated with multiple benchmark datasets for evaluation. Table~\ref{tab:datasets_results} summarizes the most widely used benchmarks along with state-of-the-art (SOTA) model performance. Further details are provided in Appendixes~\ref{sec:mpc_all_datasets} and~\ref{sec:mpc_all_results}.

These datasets vary in size, domain, and modality and are typically designed to evaluate a single MPCA skill, such as emotion recognition~\cite{poria-etal-2019-meld}, personality detection~\cite{sanchez2012nonverbal}, or dialogue summarization~\cite{zhong-etal-2021-qmsum}, which may limit their generalizability. Domain specificity also presents a challenge. For example, the AMI Meeting Corpus~\cite{carletta2005ami}, despite offering annotations for multiple tasks, is underutilized due to its narrow focus on formal product design meetings (see Table~\ref{tab:all_datasets}). 

A notable limitation is the scarcity of robust \emph{multi-modal} datasets that integrate non-verbal cues such as facial expressions, gaze, or vocal tone—essential for capturing participants' mental states and enriching semantic context.

Analysis of SOTA models across benchmarks reveals significant performance disparities. Tasks involving agent action modeling (e.g., turn detection, addressee and response selection) often exceed $80\%$ accuracy, while most other tasks average around $60\%$. Although certain models achieve near-perfect scores on specific benchmarks (e.g., Dense Retrieval models attaining 91\% recall on Ubuntu IRC~\cite{lanExploringDenseRetrieval2022,kummerfeld-etal-2019-large}), such datasets are often confined to niche domains, limiting their real-world applicability and transferability. Moreover, SOTA models for multi-modal benchmarks remain scarce, despite the critical role of cross-modal information in enhancing MPCA performance.

Overall, these observations underscore the limitations of current benchmarking practices and highlight the need for more comprehensive, diverse, and multi-modal datasets to advance future research.

\section{Future Directions}
Building intelligent MPCAs still remains an open challenge.
While our survey highlights significant progress, there remain large gaps in performance. We recommend focusing on three major research thrusts as follows to tackle these challenges.

\paragraph{Theory of Mind:} Theory of Mind (ToM) refers to the ability to understand, reason, and build beliefs about what participants might be thinking or feeling at each step of a conversation, even if those thoughts and feelings are not explicitly expressed~\cite{Premack1978}. ToM enables agents to reason and infer the beliefs and intentions of their own and other participants. 

ToM is an essential skill for MPCAs as it falls under the umbrella of \textit{State of Mind Modeling} ~\cite{vanderlyn-etal-2025-understanding}. Techniques like MindDial~\cite{qiu-etal-2024-minddial} and benchmarks like MuMA-TOM~\cite{shiMuMAToMMultimodalMultiAgent2025} and KokoMind~\cite{kokomind2023} show promise in modeling of ToM in conversations. However, the benchmarking is done on synthetic datasets, thereby not emulating realistic multi-party settings for ToM. There are works probing LLMs for ToM reasoning capabilities~\cite{chen-etal-2024-tombench, hou-etal-2024-timetom}, but studies show that scaling up LLMs does not always improve ToM reasoning~\cite{sclar-etal-2023-minding}. 
On standard ToM tasks, some LLMs achieve near-human performance~\cite{ma-etal-2023-towards-holistic}, suggesting the ability to track beliefs and intentions. 
Nevertheless, there is an ongoing debate as to whether these models truly reason or simply memorize patterns about mental states~\cite{ullman2023largelanguagemodelsfail, Kosinski_2024}. 
Latest benchmarks, ToMATO~\cite{shinoda2025tomatoverbalizingmentalstates}, reveal gaps for strong models like GPT-4.

Also, there is little work in integrating ToM capabilities into other MPCA tasks. Our intuition is that such integration will improve MPCAs performance. Table \ref{tab:model_performance} is a small experiment where we show this by finetuning two open-source models, LLaMA3.1 8B and Mistral v0.3 7B, on a popular dialogue act recognition dataset, EMOTyDA~\cite{saha-etal-2020-towards}, which also contains emotion tags of participants. We model two scenarios, one where the models predict the action based solely on the conversation till now (Act) and one where the emotions of participants are provided as additional context. The table shows that the performance of small language models improves when trained to perform dialogue act classification with emotion recognition (one capability identified in ToM) as additional context. If belief tracking (a core part of ToM) could also be integrated in a similar manner, it could lead to more intelligent MPCAs.

\begin{table}[!htb]
\centering
\resizebox{0.8\columnwidth}{!}{%
\begin{tabular}{l|l|c}
\hline
\textbf{Model} & \textbf{Task} & \textbf{F1-Score} \\
\hline\hline
\multirow{2}{*}{LLaMA3.1 8B} & Act & 55.01\% \\
& Act w. Emotion & \textbf{55.80\%} \\
\hline
\multirow{2}{*}{Mistral v0.3 7B} & Act & 43.34\% \\
& Act w. Emotion & \textbf{55.94\%} \\
\hline
\end{tabular}
}
\caption{
Results of open-source LLMs finetuned ($1$ epoch, temperature=$0.01$) on the EMOTyDA dataset for the dialogue act recognition task without and with participant emotion as additional context. We see that emotion as context improves performance in both cases, highlighting that state of mind modeling, and thereby ToM, is heavily intertwined with other MPCA tasks.
}
\label{tab:model_performance}
\vspace{-.5em}
\end{table}

To address all the gaps highlighted till now, future research should explore transforming traditional state-of-mind modeling tasks to scale ToM modeling for MPCAs, integrating emotion, persona, engagement, and intention reasoning along with belief tracking.

\paragraph{Multi-Modal Fusion and Grounding:} 

From Section \ref{sec:eval_datasets}, we see that multi-modal modeling is an underexplored area in key MPC tasks like addressee selection, conversation disentanglement, and agent action modeling. Conversation disentanglement, in particular, could benefit from integrating modalities like spatial audio and/or face tracking, but most works are still text-based. In some cases, multi-modal dialogue models with video context fail to consistently outperform text-only baselines on tasks like dialogue act recognition and response generation~\cite{wang2024friendsmmcdatasetmultimodalmultiparty}. Additionally, there is a lack of multi-modal ToM benchmarking despite promising results on synthetic datasets~\cite{shiMuMAToMMultimodalMultiAgent2025}.
Future research directions should develop methods to fuse different modalities effectively to address these identified gaps to create more robust MPCAs.

\paragraph{Evaluation Benchmarks:} 
xAs per Section \ref{sec:eval_datasets}, we observe three limitations of current MPCA evaluation benchmarks: (i) each dataset focuses on a single specific skill of MPCAs; (ii) lack of benchmarks for multi-modal settings; and (iii) lack of realistic metrics/simulations for evaluation. There have been works addressing the lack of multi-modal benchmarks~\cite{mahajan-shaikh-2021-need} through developing datasets like Friends-MMC~\cite{wang2024friendsmmcdatasetmultimodalmultiparty} for response prediction, and MuMA-TOM~\cite{shiMuMAToMMultimodalMultiAgent2025} for mental-state Q\&A. However, they are mostly synthetic data covering individual tasks, thereby not solving the first. Furthermore, most evaluation is done using static metrics like BLEU or accuracy, which do not fully capture the performance of MPCAs' interactions. 
Future research must create real-world multi-modal benchmarks with simulation-based metrics where an MPCA is inserted into multi-party settings to interact in real time~\cite{gunsonHolisticEvaluation2024}.


\section{Conclusion}
The development of multi-party conversation agents has come a long way from text-based machine learning to multi-modal and LLM-based approaches. In this survey paper, we discussed the recent literature on MPCAs by defining a thematic taxonomy and categorizing works/tasks according to it. Our survey provides details on how different methods solve each problem of MPCAs. It also provides insights into the limitations and challenges faced when developing MPCAs. Finally, we discuss future directions for this field of research, specifically how Theory of Mind, multi-modal fusion, and grounding, and better benchmarks can help us develop better MPCAs. Addressing these aspects will advance MPCAs toward genuinely engaging and socially aware interactions, bridging the gap between current systems and human-like conversational intelligence.



\section*{Limitations}
Our survey primarily focuses on research published after 2021, which may exclude some foundational or emerging contributions outside this period. Given the extensive body of related work, we acknowledge the possibility of overlooking equally significant studies. The future directions we propose are based on our interpretation of current research trends and supported by cursory experiments, which may not capture all perspectives. Additionally, many works we reviewed rely on custom datasets and models, some of which do not align with recent standard benchmarks. This reliance may introduce biases in our analysis. We recommend further, more comprehensive research to cross-verify our findings.


\bibliography{anthology,custom,references}

\appendix
\section{MPC Thematic Taxonomy}
\label{sec:mpc_task_taxonomy_full}
\tikzstyle{my-box}=[
    rectangle,
    draw=hidden-draw,
    rounded corners,
    text opacity=1,
    minimum height=1.5em,
    minimum width=5em,
    inner sep=2pt,
    align=center,
    fill opacity=.5,
    line width=0.8pt,
]
\tikzstyle{leaf}=[my-box, minimum height=1.5em,
    fill=hidden-pink!80, text=black, align=left,font=\normalsize,
    inner xsep=2pt,
    inner ysep=4pt,
    line width=0.8pt,
]
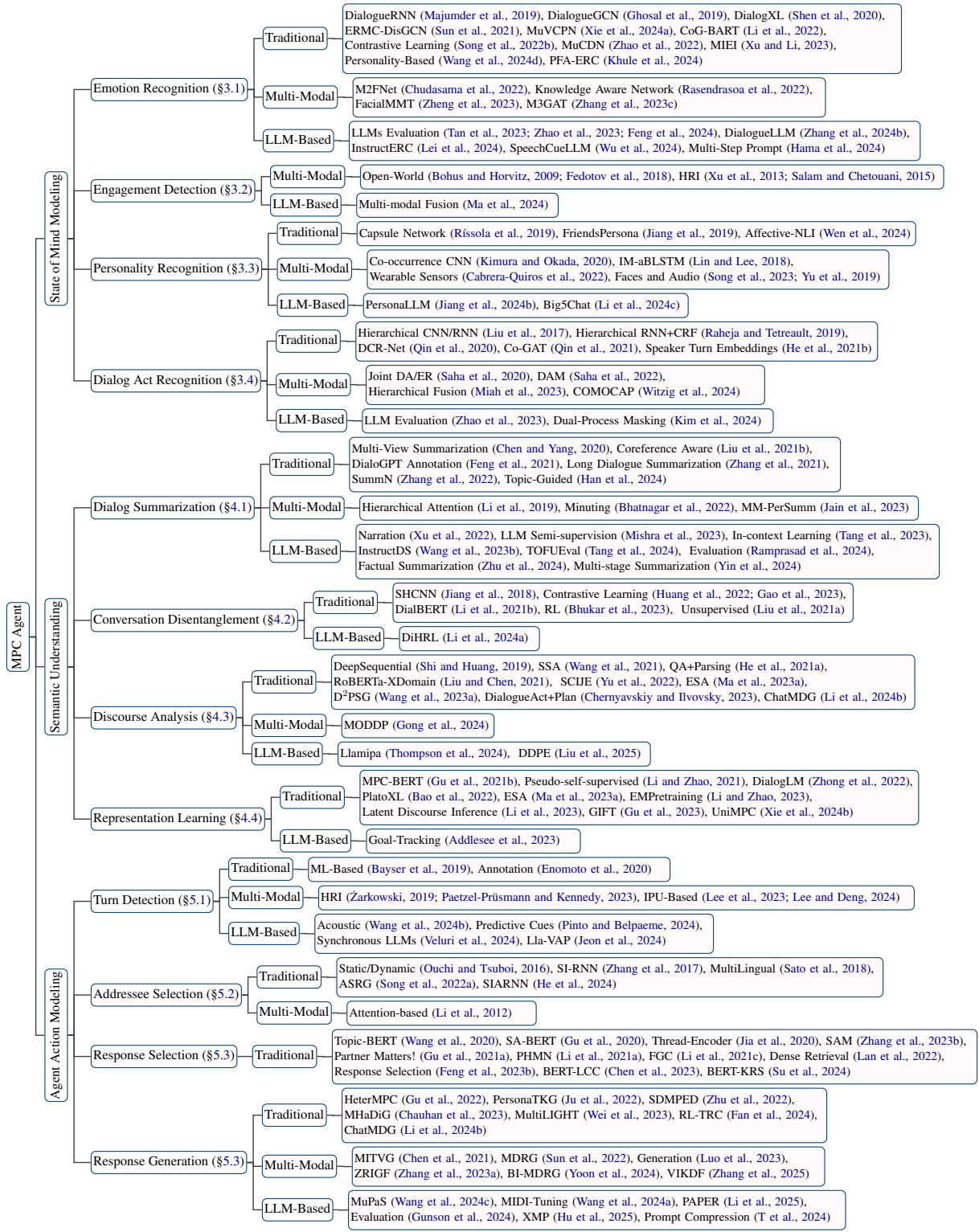
\begin{figure*}[!htb]
    \vspace{-1.0cm}
    \centering
    \resizebox{\textwidth}{!}{
        \begin{forest}
            forked edges,
            for tree={
                grow=east,
                reversed=false,
                anchor=base west,
                parent anchor=east,
                child anchor=west,
                base=left,
                font=\large,
                rectangle,
                draw=hidden-draw,
                rounded corners,
                align=left,
                minimum width=4em,
                edge+={darkgray, line width=1pt},
                s sep=3pt,
                inner xsep=2pt,
                inner ysep=3pt,
                line width=0.8pt,
                ver/.style={rotate=90, child anchor=north, parent anchor=south, anchor=center},
            },
            [
                MPC Agent, ver
                [
                    Agent Action Modeling, ver
                    [
                        Response Generation (\S \ref{sec:response_selection_generation})
                        [
                            LLM-Based
                            [
                                MuPaS~\cite{wangMultiPartySupervisedFinetuning2024}{, }MIDI-Tuning~\cite{wangInstructOnceChat2024}{, }PAPER~\cite{liPAPERPersonaAwareChainofThought2025}{, }\\Evaluation~\cite{gunsonHolisticEvaluation2024}{, }XMP~\cite{huCanXLLMsUnderstand2025}{, }Prompt Compression~\cite{tThreadDetectionResponse2024a}, leaf, text width=40.0em
                            ]
                        ]
                        [
                            Multi-Modal
                            [
                                MITVG~\cite{chenMultimodalIncrementalTransformer2021}{, }MDRG~\cite{sunMultimodalDialogueResponse2022}{, }Generation~\cite{luoForwardCreationReverse2023}{, }\\ZRIGF~\cite{zhangZRIGFInnovativeMultimodal2023}{, }BI-MDRG~\cite{yoonBIMDRGBridgingImage2024}{, }VIKDF~\cite{zhangDistillingImplicitMultimodal2025}, leaf, text width=38.0em
                            ]
                        ]
                        [
                            Traditional
                            [
                                HeterMPC~\cite{guHeterMPCHeterogeneousGraph2022}{, }PersonaTKG~\cite{juLearningImprovePersona2022}{, }SDMPED~\cite{zhuMultiPartyEmpatheticDialogue2022}{, }\\MHaDiG~\cite{chauhanMHaDiGMultilingualHumoraided2023}{, }MultiLIGHT~\cite{weiMultiPartyChatConversational2023a}{, }RL-TRC~\cite{fanImprovingMultipartyDialogue2024}{, }\\ChatMDG~\cite{liChatMDGDiscourseParsing2024}, leaf, text width=39.0em
                            ]
                        ]
                    ]
                    [
                        Response Selection (\S \ref{sec:response_selection_generation})
                        [
                            Traditional
                            [
                                Topic-BERT~\cite{wangResponseSelectionMultiParty2020}{, }SA-BERT~\cite{guSpeakerAwareBERTMultiTurn2020}{, }Thread-Encoder~\cite{jiaMultiturnResponseSelection2020}{, }SAM~\cite{zhangSAMMultiturnResponse2023}{, }\\Partner Matters!~\cite{guPartnerMattersEmpirical2021}{, }PHMN~\cite{liDialogueHistoryMatters2021}{, }FGC~\cite{liSmallChangesMake2021}{, }Dense Retrieval~\cite{lanExploringDenseRetrieval2022}{, }\\Response Selection~\cite{fengLearningMultiturnResponse2023}{, }BERT-LCC~\cite{chenImprovingBERTLocal2023}{, }BERT-KRS~\cite{suBERTKRSBERTBasedModel2024}, leaf, text width=52em
                            ]
                        ]
                    ]
                    [
                        Addressee Selection (\S \ref{sec:addressee_selection})
                        [
                            Multi-Modal
                            [
                                Attention-based~\cite{li2012attnaddresseeselect}, leaf, text width=15.0em
                            ]
                        ]
                        [
                            Traditional
                            [
                                Static/Dynamic~\cite{ouchi2016addresseerespselect}{, }SI-RNN~\cite{zhang2017addrespselrnn}{, }MultiLingual~\cite{sato2018multilingualaddrsel}{, }\\ASRG~\cite{song2022jointasrg}{, }SIARNN~\cite{he2024addrselmpccyber}, leaf, text width=44.0em
                            ]
                        ]
                    ]
                    [
                        Turn Detection (\S \ref{sec:turn_detect})
                        [
                            LLM-Based
                            [
                                Acoustic~\cite{wang2024turnacousticllm}{, }Predictive Cues~\cite{pinto2024predictiveturn}{, }\\Synchronous LLMs~\cite{veluri2024syncllmduplexagents}{, }Lla-VAP~\cite{jeon2024llavap}, leaf, text width=32.0em
                            ]
                        ]
                        [
                            Multi-Modal
                            [
                                HRI~\cite{zarkowski2019mpxturntakinghri,paetzel2023robotturntaking}{, }IPU-Based~\cite{lee2023turnanalysismpc, lee2024endofturn}, leaf, text width=48.0em
                            ]
                        ]
                        [
                            Traditional
                            [
                                ML-Based~\cite{bayser2019mlmpcdlogs}{, }Annotation~\cite{enomoto2020analyticturnannot}, leaf, text width=29.0em
                            ]
                        ]
                    ]
                ]    
                [
                    Semantic Understanding, ver
                    [
                        Representation Learning (\S \ref{sec:repr_learning})
                        [
                            LLM-Based
                            [
                                Goal-Tracking~\cite{addlesee2023mpcgoaltrackllm}, leaf, text width=17.0em
                            ]
                        ]
                        [
                            Traditional
                            [
                                MPC-BERT~\cite{gu2021mpcbert}{, }Pseudo-self-supervised~\cite{li2021selfpseudosupervisedmpc}{, }DialogLM~\cite{zhong2022dialoglm}{, }\\PlatoXL~\cite{bao2022platoxl}{, }ESA~\cite{ma2023enhanced}{, }EMPretraining~\cite{li2023empretainingmpdrg}{, }\\Latent Discourse Inference~\cite{li2023latentdiscourseinf}{, }GIFT~\cite{gu-etal-2023-gift}{, }UniMPC~\cite{xie2024unimpc}, leaf, text width=45.0em
                            ]
                        ]
                    ]
                    [
                        Discourse Analysis (\S \ref{sec:discourse_structure})
                        [
                            LLM-Based
                            [
                                Llamipa~\cite{thompsonLlamipaIncrementalDiscourse2024}{, } DDPE~\cite{liuEnhancingMultipartyDialogue2025}, leaf, text width=25.0em
                            ]
                        ]
                        [
                            Multi-Modal
                            [
                                MODDP~\cite{gongMODDPMultimodalOpendomain2024}, leaf, text width=12.0em
                            ]
                        ]
                        [
                            Traditional
                            [
                                DeepSequential~\cite{shiDeepSequentialModel2019}{, }SSA~\cite{wangStructureSelfAwareModel2021}{, }QA+Parsing~\cite{heMultitaskingDialogueComprehension2021}{, }\\RoBERTa-XDomain~\cite{liuImprovingMultiPartyDialogue2021}{, } SCIJE~\cite{yuSpeakerAwareDiscourseParsing2022}{, }ESA~\cite{ma2023enhanced}{, }\\D$^2$PSG~\cite{wangD2PSGMultiPartyDialogue2023}{, }DialogueAct+Plan~\cite{chernyavskiyTransformerbasedMultiPartyConversation2023a}{, }ChatMDG~\cite{liChatMDGDiscourseParsing2024}, leaf, text width=47.0em
                            ]
                        ]
                    ]
                    [
                        Conversation Disentanglement (\S \ref{sec:conv_disentangle})
                        [
                            LLM-Based
                            [
                                DiHRL~\cite{li2024dialogdisentangle}, leaf, text width=11.0em
                            ]
                        ]
                        [
                            Traditional
                            [
                                SHCNN~\cite{jiangLearningDisentangleInterleaved2018}{, }Contrastive Learning~\cite{huangConversationDisentanglementBiLevel2022,gaoCluCDDContrastiveDialogue2023}{, }\\DialBERT~\cite{liDialBERTHierarchicalPreTrained2021}{, }RL~\cite{bhukarEndtoEndDeepReinforcement2023}{, } Unsupervised~\cite{liuUnsupervisedConversationDisentanglement2021}, leaf, text width=37.0em
                            ]
                        ]
                    ]
                    [
                        Dialog Summarization (\S \ref{sec:dialog_summ})
                        [
                            LLM-Based
                            [
                                Narration~\cite{xu2022narratedialog}{, }LLM Semi-supervision~\cite{mishra2023llmaidedextractsumm}{, }In-context Learning~\cite{tang2023iclllmdialogsumm}{, }\\InstructDS~\cite{wang2023instructds}{, }TOFUEval~\cite{tang2024tofueval}{, } Evaluation~\cite{ramprasad2024llmbehavinds}{, }\\Factual Summarization~\cite{zhu2024factualdialogsummfromllm}{, }Multi-stage Summarization~\cite{yin2024twostagellmdialogsumm}, leaf, text width=47.0em
                            ]
                        ]
                        [
                            Multi-Modal
                            [
                                Hierarchical Attention~\cite{li_keep_on_topic_2019}{, }Minuting~\cite{bhatnagar2022multilingualmpcminute}{, }MM-PerSumm~\cite{jain2023multimodalperspectivesumm}, leaf, text width=45.0em
                            ]
                        ]
                        [
                            Traditional
                            [
                                Multi-View Summarization~\cite{chen2020multiviewseq2seqds}{, }Coreference Aware~\cite{liu2021corefawareds}{, }\\DialoGPT Annotation~\cite{feng2021dilogptannotatedsummarizer}{, }Long Dialogue Summarization~\cite{zhang2021long_dialog_summ}{, }\\SummN~\cite{zhang2022summn}{, }Topic-Guided~\cite{han2024tgds}, leaf, text width=39.5em
                            ]
                        ]
                    ]
                ]
                [
                    State of Mind Modeling, ver
                    [
                        Dialog Act Recognition (\S \ref{sec:dar})
                        [
                            LLM-Based
                            [
                                LLM Evaluation~\cite{zhao2023chatgptequippedemotionaldialogue}{, }Dual-Process Masking~\cite{kim-etal-2024-dual}, leaf, text width=33.0em
                            ]
                        ]
                        [
                            Multi-Modal
                            [
                                Joint DA/ER~\cite{saha-etal-2020-towards}{, }DAM~\cite{saha-etal-2022-meta}{, }\\Hierarchical Fusion~\cite{miah-etal-2023-hierarchical}{, }COMOCAP~\cite{witzig2024multimodaldac}, leaf, text width=32.0em
                            ]
                        ]
                        [
                            Traditional
                            [
                                Hierarchical CNN/RNN~\cite{liu-etal-2017-using-context}{, }Hierarchical RNN+CRF~\cite{raheja-tetreault-2019-dialogue}{, }\\DCR-Net~\cite{Qin_Che_Li_Ni_Liu_2020}{, }Co-GAT~\cite{Qin_Li_Che_Ni_Liu_2021}{, }Speaker Turn Embeddings~\cite{he-etal-2021-speaker-turn}, leaf, text width=42.0em
                            ]
                        ]
                    ]
                    [
                        Personality Recognition (\S \ref{sec:personality_recognition})
                        [
                            LLM-Based
                            [
                                PersonaLLM~\cite{jiang-etal-2024-personallm}{, }Big5Chat~\cite{li2024big5chatshapingllmpersonalities}, leaf, text width=26.0em
                            ]
                        ]
                        [
                            Multi-Modal
                            [
                                Co-occurrence CNN~\cite{kimura2020cooccurrencecnn}{, }IM-aBLSTM~\cite{lin2018interlocutormodattentblstm}{, }\\Wearable Sensors~\cite{quiros2022personalitywearabledevcam}{, }Faces and Audio~\cite{song2023personspecificcognition,yu2019identifyingpersonalitytraitsusing}, leaf, text width=42.0em
                            ]
                        ]
                        [
                            Traditional
                            [
                                Capsule Network~\cite{rissola2019prccapsnet}{, }FriendsPersona~\cite{jiang2019automatictextbasedpersonalityrecognition}{, }Affective-NLI~\cite{wen2024affectivenli}, leaf, text width=45.0em
                            ]
                        ]
                    ]
                    [
                        Engagement Detection (\S \ref{sec:engagement_detection})
                        [
                            LLM-Based
                            [
                                Multi-modal Fusion~\cite{maMultimodalFusionLLMs2024}, leaf, text width=16.0em
                            ]
                        ]
                        [
                            Multi-Modal
                            [
                                Open-World~\cite{bohusModelsMultipartyEngagement2009,fedotovMultimodalApproachEngagement2018}{, }HRI~\cite{xuDesigningEngagementawareAgents2013,salamEngagementDetectionBased2015}, leaf, text width=47.0em
                            ]
                        ]
                    ]
                    [
                        Emotion Recognition (\S \ref{sec:emotion_recognition})
                        [
                            LLM-Based
                            [
                                LLMs Evaluation~\cite{tan-etal-2023-chatgpt,zhao2023chatgptequippedemotionaldialogue,feng-etal-2024-affect}{, }DialogueLLM~\cite{zhang2024dialoguellmcontextemotionknowledgetuned}{, }\\InstructERC~\cite{lei2024instructercreformingemotionrecognition}{, }SpeechCueLLM~\cite{wu2024silentlettersamplifyingllms}{, }Multi-Step Prompt~\cite{hama2024ercmultistepprompt}, leaf, text width=46.0em
                            ]
                        ]
                        [
                            Multi-Modal
                            [
                                M2FNet~\cite{chudasama2022m2fnet}{, }Knowledge Aware Network~\cite{rasendrasoa2022realtimemultimodalerc}{, }\\FacialMMT~\cite{zheng-etal-2023-facial}{, }M3GAT~\cite{zhang2023m3gat}, leaf, text width=38.0em
                            ]
                        ]
                        [
                            Traditional
                            [
                                DialogueRNN~\cite{Majumder_Poria_Hazarika_Mihalcea_Gelbukh_Cambria_2019}{, }DialogueGCN~\cite{ghosal-etal-2019-dialoguegcn}{, }DialogXL~\cite{shen2020dialogxlallinonexlnetmultiparty}{, }\\ERMC-DisGCN~\cite{sun-etal-2021-discourse-aware}{, }MuVCPN~\cite{Xie2024MultiViewCP}{, }CoG-BART~\cite{Li_Yan_Qiu_2022}{, }\\Contrastive Learning~\cite{song-etal-2022-supervised}{, }MuCDN~\cite{zhao-etal-2022-mucdn}{, }MIEI~\cite{xu2023miei}{, }\\Personality-Based~\cite{wang-etal-2024-emotion}{, }PFA-ERC~\cite{khule-etal-2024-pfa}, leaf, text width=45.0em
                            ]
                        ]
                    ]
                ]  
            ]
        \end{forest}
    }
    \caption{Full taxonomy of MPC tasks and recent work under them.
    }
    \label{fig:full_taxo_of_mpc_tasks}
\end{figure*}
To complement the brief taxonomy presented in Figure~\ref{fig:taxo_of_mpc_tasks}, we provide a comprehensive breakdown of the multi-party conversational studies in Figure~\ref{fig:full_taxo_of_mpc_tasks}. We expand each of the three themes—State of Mind Modeling, Semantic Understanding, and Agent Action Modeling—by listing all associated tasks and surveying representative works under each category. Further, we organize the literature based on the methodology adopted: traditional approaches, multi-modal systems, and LLM-based techniques. This detailed taxonomy serves as a valuable reference for understanding the full scope of MPC research and identifying key trends across different modeling strategies.

\section{MPC Datasets}
\label{sec:mpc_all_datasets}
In this section, we provide a comprehensive list of multi-party conversational (MPC) datasets we reviewed through our survey. We expand the datasets in Table~\ref{tab:all_datasets} to include a broader range of datasets spanning diverse domains, modalities, and annotated task types. This detailed compilation supports further exploration and benchmarking across the different MPCA skill tasks discussed in this paper.

\begin{table*}[!htb]\large
  \centering
  \resizebox{\textwidth}{!}{%
    \begin{filecontents*}{data/all_datasets.tsv}
Dataset	Tasks	Modality	Description	
MELD \cite{poria-etal-2019-meld}	Emotion Recognition	Audio, Video, Text	1,400 dialogues and approximately 13,000 utterances extracted from the TV series Friends	
EmotionLines \cite{hsu-etal-2018-emotionlines}	Emotion Recognition	Text	 2,000 dialogues with a total of 29,245 utterances	
Akan Cinematic Emotions (ACE) \cite{sasu2025akancinematicemotionsace}	Emotion Recognition	Audio, Video, Text	385 dialogues and 6,162 utterances sourced from 21 Akan-language movies	
EmoryNLP \cite{zahiri2017emotiondetectiontvtranscripts}	Emotion Recognition, Personality Recognition	Text	897 conversations, 12606 utterances from the TV series Friends	
EMOTyDA \cite{saha-etal-2020-towards}	Emotion Recognition, Dialog Act Recognition	Audio, Video, Text	Annotated subset of MELD and IEMOCAP	
DailyDialog \cite{li-etal-2017-dailydialog}	Emotion Recognition, Dialog Act Recognition	Text	13118 conversations, 102979 utterances crawled from English learning websites	
AMI Meeting Corpus \cite{carletta2005ami}	Emotion Recognition, Personality Recognition, Dialog Act Recognition, Dialogue Summarization	Audio, Video, Text	100 hours of meeting recordings	
MPIIGroupInteraction \cite{muller2024mpiigroupinteraction}	Engagement Detection	Audio, Video	22 group discussions each with 3 to 4 participants exceeding 440 minutes of total recordings	
Emergent LEAder (ELEA) \cite{sanchez2012nonverbal}	Personality Recognition	Audio, Video	27 meeting recordings each with 3 to 4 participants performing winter survival task	
Team Corpus \cite{litman-etal-2016-teams}	Personality Recognition	Audio	47 hours of multiparty interaction from 62 teams (3 to 4 person)  playing the collaborative board game Forbidden Island	
MRDA Corpus \cite{shriberg-etal-2004-icsi}	Dialog Act Recognition	Audio, Text	75 meetings with an average of about six participants per meeting	
MIntRec2.0 \cite{zhang2024mintrec20largescalebenchmarkdataset}	Dialog Act Recognition	Audio, Video, Text	1,245 conversations, 15,040 utterances from TV series: Superstore, The Big Bang Theory, and Friends	
TopDial \cite{wang-etal-2023-target}	Dialog Act Recognition	Text	18,000 dialogues with 12.3 average utterances per dialogue	
QMSum \cite{zhong-etal-2021-qmsum}	Dialogue Summarization	Text	1,808 query-summary pairs derived from 232 meetings across various domains	
SAMSum \cite{gliwa-etal-2019-samsum}	Dialogue Summarization	Text	16,000 messenger-like conversations	
MDS \cite{liu-etal-2024-mds}	Dialogue Summarization	Audio, Video, Text	16,000 minutes of video clips, with dialogues from TV series such as "The Big Bang Theory" and TikTok videos	
Ubuntu IRC \cite{kummerfeld-etal-2019-large}	Conversation Disentanglement, Addressee Selection	Text	19,560 conversations collected from Ubuntu Internet Relay Chat (IRC) logs	
Movie Dialogue Dataset \cite{chang-etal-2023-dramatic}	Conversation Disentanglement	Text	2209 conversational threads derived from scripts of 831 movies and TV series	
Multi-Domain Chat Disentanglement Dataset \cite{li2024dialogdisentangle}	Conversation Disentanglement	Text	875 conversations across 4 distinct IRC channels	
STAC \cite{asher-etal-2016-discourse}	Discourse Parsing	Text	~1200 dialogues from from an online game (Settlers of Catan)	
Molweni \cite{li-etal-2020-molweni}	Discourse Parsing	Text	10,000 dialogues built upon the multi-party subset of the Ubuntu Dialogue Corpus	
MODDP \cite{gongMODDPMultimodalOpendomain2024}	Discourse Parsing	Audio, Video, Text	864 dialogues, 18,114 utterances Chinese dataset with 12.7 hours of video clips	
MMDialog \cite{feng-etal-2023-mmdialog}	General	Text + Image	1.4 million conversations sourced from youtube comment threads with timed video snapshot
\end{filecontents*}
\csvreader[
      separator=tab,
      tabular=p{5cm}|p{5cm}|p{3cm}|p{10cm},
      table head=\hline \textbf{Dataset} & \textbf{Task} & \textbf{Modality} & \textbf{Description} \\ \hline\hline,
      late after line=\\\hline
    ]{data/all_datasets.tsv}{}%
    {\csvcoli & \csvcolii & \csvcoliii & \csvcoliv}
  }
  \caption{Extended list of multi-party conversational datasets
  \label{tab:all_datasets}
  }
\end{table*}

\section{Results}
\label{sec:mpc_all_results}
Table~\ref{tab:all_results} provides a comprehensive overview of model performances across a wide range of MPCA tasks and datasets. This table captures the broader landscape of research efforts, including emerging baselines, multi-modal models, and task-specific strategies. Notably, we observe significant performance gaps across tasks like engagement detection, discourse parsing, and multi-modal response generation, underscoring persistent modeling challenges. Additionally, it reveals increased interest in multi-modal approaches, yet most still underperform compared to unimodal text-based models, highlighting the need for more robust multi-modal integration in future MPCA studies.
\begin{table*}[!htb]\Large
    \centering
    \resizebox{\textwidth}{!}{%
    \begin{tabular}{l|c|l|c|c}
    \hline
        \textbf{Model} & \textbf{Task} & \textbf{Dataset} & \textbf{Eval Metric} & \textbf{Eval Result} \\ \hline\hline
        DialogueLLM \cite{zhang2024dialoguellmcontextemotionknowledgetuned} & Emotion Recognition & EmoryNLP \cite{zahiri2017emotiondetectiontvtranscripts} & F1 Score & 40.05 \\ \hline
        DialogueLLM \cite{zhang2024dialoguellmcontextemotionknowledgetuned} & Emotion Recognition & MELD \cite{poria-etal-2019-meld} & F1 Score & 71.9 \\ \hline
        InstructERC \cite{lei2024instructercreformingemotionrecognition} & Emotion Recognition & EmoryNLP \cite{zahiri2017emotiondetectiontvtranscripts} & F1 Score & 41.37 \\ \hline
        InstructERC \cite{lei2024instructercreformingemotionrecognition} & Emotion Recognition & MELD \cite{poria-etal-2019-meld} & F1 Score & 69.15 \\ \hline
        M3GAT \cite{zhang2023m3gat} & Emotion Recognition & MELD \cite{poria-etal-2019-meld} & F1 Score & 40.53 \\ \hline
        PFA-ERC \cite{khule-etal-2024-pfa} & Emotion Recognition & EmoryNLP \cite{zahiri2017emotiondetectiontvtranscripts} & F1 Score & 42.94 \\ \hline
        PFA-ERC \cite{khule-etal-2024-pfa} & Emotion Recognition & MELD \cite{poria-etal-2019-meld} & F1 Score & 68.73 \\ \hline
        SpeechCueLLM \cite{wu2024silentlettersamplifyingllms} & Emotion Recognition & MELD \cite{poria-etal-2019-meld} & F1 Score & 67.6 \\ \hline
        Multi-Modal Fusion \cite{maMultimodalFusionLLMs2024} & Engagement Detection & Custom & RMSE & 1.338 \\ \hline
        Affect-NLI \cite{wen2024affectivenli} & Personality Recognition & EmoryNLP \cite{zahiri2017emotiondetectiontvtranscripts} & Accuracy & 61.3 \\ \hline
        ChatGPT \cite{zhao2023chatgptequippedemotionaldialogue} & Dialog Act & DailyDialog \cite{li-etal-2017-dailydialog} & F1 Score & 70 \\ \hline
        COMOCAP \cite{witzig2024multimodaldac} & Dialog Act & DailyDialog \cite{li-etal-2017-dailydialog} & F1 Score & 79.66 \\ \hline
        DAM (Text+Video) \cite{saha-etal-2022-meta} & Dialog Act & EMOTyDA \cite{saha-etal-2020-towards} & F1 Score & 57.01 \\ \hline
        Hierarchical Fusion \cite{miah-etal-2023-hierarchical} & Dialog Act & EMOTyDA \cite{saha-etal-2020-towards} & F1 Score & 50.3 \\ \hline
        Hierarchical Fusion \cite{miah-etal-2023-hierarchical} & Dialog Act & MRDA Corpus \cite{shriberg-etal-2004-icsi} & F1 Score & 29.11 \\ \hline
        In-Context Learning \cite{tang2023iclllmdialogsumm} & Dialog Summary & SAMSum \cite{gliwa-etal-2019-samsum} & ROUGE-1 & 34.6 \\ \hline
        InstructDS \cite{wang2023instructds} & Dialog Summary & SAMSum \cite{gliwa-etal-2019-samsum} & ROUGE-1 & 58.4 \\ \hline
        Minuting \cite{bhatnagar2022multilingualmpcminute} & Dialog Summary & SAMSum \cite{gliwa-etal-2019-samsum} & ROUGE-1 & 45 \\ \hline
        Bi-Level CL \cite{huangConversationDisentanglementBiLevel2022} & Disentanglement & Ubuntu IRC \cite{kummerfeld-etal-2019-large} & F1 Score & 70.4 \\ \hline
        CLUCDD \cite{gaoCluCDDContrastiveDialogue2023} & Disentanglement & Ubuntu IRC \cite{kummerfeld-etal-2019-large} & F1 Score & 58.42 \\ \hline
        DiHRL \cite{li2024dialogdisentangle} & Disentanglement & Ubuntu IRC \cite{kummerfeld-etal-2019-large} & F1 Score & 47.63 \\ \hline
        RL \cite{bhukarEndtoEndDeepReinforcement2023} & Disentanglement & Ubuntu IRC \cite{kummerfeld-etal-2019-large} & F1 Score & 51.9 \\ \hline
        D2PSG \cite{wangD2PSGMultiPartyDialogue2023} & Discourse Parsing & STAC \cite{asher-etal-2016-discourse} & F1 Score & 62.77 \\ \hline
        Deep Sequential \cite{shiDeepSequentialModel2019} & Discourse Parsing & STAC \cite{asher-etal-2016-discourse} & F1 Score & 55.7 \\ \hline
        Llamipa \cite{thompsonLlamipaIncrementalDiscourse2024} & Discourse Parsing & STAC \cite{asher-etal-2016-discourse} & F1 Score & 60.7 \\ \hline
        RoBERTa-XDomain \cite{liuImprovingMultiPartyDialogue2021} & Discourse Parsing & STAC \cite{asher-etal-2016-discourse} & F1 Score & 57.1 \\ \hline
        SCIJE \cite{yuSpeakerAwareDiscourseParsing2022} & Discourse Parsing & STAC \cite{asher-etal-2016-discourse} & F1 Score & 57.4 \\ \hline
        IPU-Based \cite{lee2024endofturn} & Turn Detection & Custom & F1 Score & 87.3 \\ \hline
        LLa-VAP \cite{jeon2024llavap} & Turn Detection & CCPE \cite{radlinski-etal-2019-coached} & F1 Score & 83.13 \\ \hline
        Predictive Cues \cite{pinto2024predictiveturn} & Turn Detection & DailyDialog \cite{li-etal-2017-dailydialog} & F1 Score & 90.3 \\ \hline
        ASRG \cite{song2022jointasrg} & Addressee Selection & Ubuntu IRC \cite{kummerfeld-etal-2019-large} & Accuracy & 84.65 \\ \hline
        SI-RNN \cite{zhang2017addrespselrnn} & Addressee Selection & Ubuntu IRC \cite{kummerfeld-etal-2019-large} & Accuracy & 80.47 \\ \hline
        BERT-KRS \cite{suBERTKRSBERTBasedModel2024} & Response Selection & Persona-Chat \cite{zhang-etal-2018-personalizing} & Recall & 82 \\ \hline
        BERT-LCC \cite{chenImprovingBERTLocal2023} & Response Selection & Ubuntu IRC \cite{kummerfeld-etal-2019-large} & Recall & 88.9 \\ \hline
        Dense Retrieval \cite{lanExploringDenseRetrieval2022} & Response Selection & Ubuntu IRC \cite{kummerfeld-etal-2019-large} & Recall & 91 \\ \hline
        FGC \cite{liSmallChangesMake2021} & Response Selection & Ubuntu IRC \cite{kummerfeld-etal-2019-large} & Recall & 88.6 \\ \hline
        Partner Matters! \cite{guPartnerMattersEmpirical2021} & Response Selection & Persona-Chat \cite{zhang-etal-2018-personalizing} & Recall & 70.7 \\ \hline
        PLMKS+PLMCS+PLMRanker \cite{fengLearningMultiturnResponse2023} & Response Selection & WoW \cite{dinan2019wizardwikipediaknowledgepoweredconversational} & Recall & 92.7 \\ \hline
        SA-BERT \cite{guSpeakerAwareBERTMultiTurn2020} & Response Selection & Ubuntu IRC \cite{kummerfeld-etal-2019-large} & Recall & 85.5 \\ \hline
        BI-MDRG \cite{yoonBIMDRGBridgingImage2024} & Response Generation & MMDialog \cite{feng-etal-2023-mmdialog} & BLEU & 27.6 \\ \hline
        HeterMPC \cite{guHeterMPCHeterogeneousGraph2022} & Response Generation & Ubuntu IRC \cite{kummerfeld-etal-2019-large} & BLEU & 12.61 \\ \hline
        MIDI-Tuning \cite{wangInstructOnceChat2024} & Response Generation & TopDial \cite{wang-etal-2023-target} & BLEU & 39.6 \\ \hline
        MuPaS \cite{wangMultiPartySupervisedFinetuning2024} & Response Generation & Custom & Human & 8.16/10 \\ \hline
        PAPER \cite{liPAPERPersonaAwareChainofThought2025} & Response Generation & HLA-Chat++ \cite{Li_2020_hla} & BLEU & 44.16 \\ \hline
        VIKDF \cite{zhangDistillingImplicitMultimodal2025} & Response Generation & Reddit \cite{yang2021open} & BLEU & 16.47 \\ \hline
        ZRIGF \cite{zhangZRIGFInnovativeMultimodal2023} & Response Generation & Reddit \cite{dziri-etal-2019-augmenting} & BLEU & 16.06 \\ \hline
    \end{tabular}
    }
    \caption{Performance of MPCA models on various tasks}
    \label{tab:all_results}
\end{table*}

\end{document}